\documentclass[letterpaper, 10 pt, conference]{ieeeconf}  

\usepackage{graphics} 
\usepackage{amsmath} 
\usepackage{amssymb}  

\usepackage{graphicx}
\usepackage{hyperref}
\usepackage{authblk}
\usepackage{subcaption}

\title{\LARGE \bf IMLS-SLAM: scan-to-model matching based on 3D data}

\author[1]{Jean-Emmanuel Deschaud}
\affil[1]{MINES ParisTech, PSL Research University, Centre for Robotics, 60 Bd St Michel 75006 Paris, France}

\begin{document}

\maketitle
\thispagestyle{empty}
\pagestyle{empty}

\begin{abstract}

The Simultaneous Localization And Mapping (SLAM) problem has been well studied in the robotics community, especially using mono, stereo cameras or depth sensors. 3D depth sensors, such as Velodyne LiDAR, have proved in the last 10 years to be very useful to perceive the environment in autonomous driving, but few methods exist that directly use these 3D data for odometry. We present a new low-drift SLAM algorithm based only on 3D LiDAR data. Our method relies on a scan-to-model matching framework. We first have a specific sampling strategy based on the LiDAR scans. We then define our model as the previous localized LiDAR sweeps and use the Implicit Moving Least Squares (IMLS) surface representation. We show experiments with the Velodyne HDL32 with only 0.40\% drift over a $4~km$ acquisition without any loop closure (i.e., $16~m$ drift after $4~km$). We tested our solution on the KITTI benchmark with a Velodyne HDL64 and ranked among the best methods (against mono, stereo and LiDAR methods) with a global drift of only 0.69\%.

\end{abstract}

\section{Introduction}

The localization of a vehicle is an important task in the field of autonomous driving. The current trend in research is to find solutions using accurate maps. However, when such maps are not available (an area is not mapped or there have been big changes since the last update), we need Simultaneous Localization And Mapping (SLAM) solutions. There are many such solutions based on different sensors, such as cameras (mono or stereovision), odometers and depth sensors or a combination of these sensors.

The advantage of LiDARs with respect to cameras is that the noise associated with each distance measurement is independent of the distance and the lighting conditions. However, the amount of data to process and the sparse density of collected range images are still challenging. In this paper, we present a new scan-to-model framework using an implicit surface representation of the map inspired by previous RGB-D methods to better handle the large amount and sparsity of acquired data. The result is low-drift LiDAR odometry and an improvement in the quality of the mapping.

\begin{figure}[!ht]
\begin{center}
\includegraphics[width=0.92\linewidth]{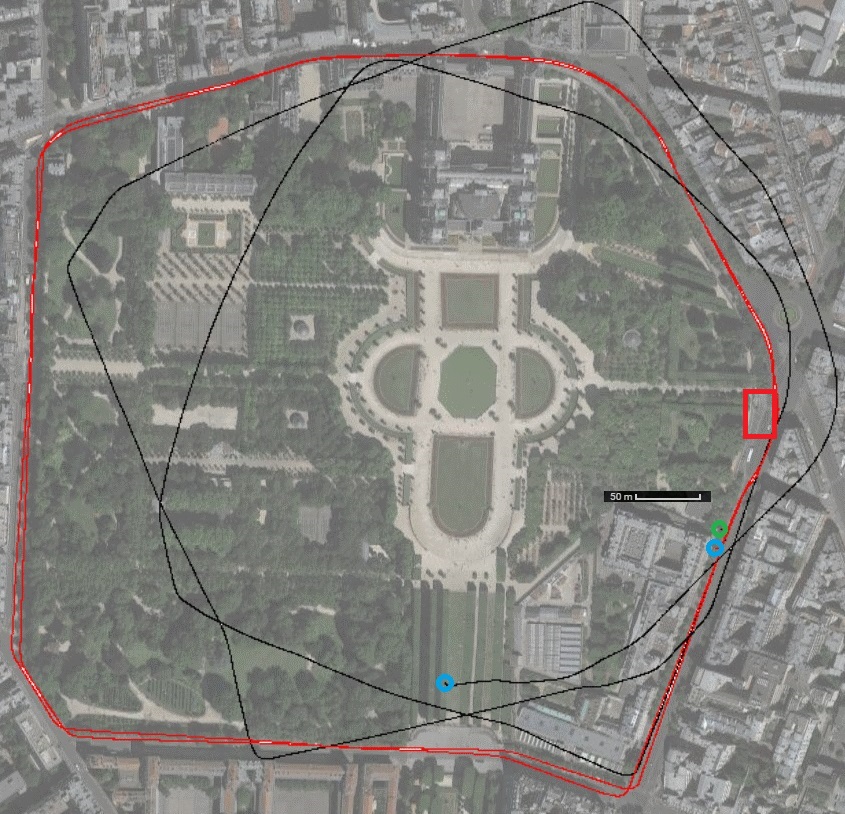}
\end{center}
\caption[width=\linewidth]{Trajectory in red of our IMLS SLAM with a vertical Velodyne HDL32 on top of a car (two loops of $2~km$ length) in the city of Paris. We can see the good superposition of the two loops. In black is the trajectory of the classical scan-to-scan matching. The starting point of both methods is the same green circle. The end point of the IMLS SLAM and the end point of the scan-to-scan matching are inside the blue circles.}
\label{fig:traj_velo32_paris}
\end{figure}

\section{Related Work}

There are hundreds of works on SLAM in the literature. Here, we only present the recent advances in six degrees of freedom (6-DOF) SLAM LiDAR with 3D mapping. Most LiDAR approaches are variations of the traditionnal iterative closest point (ICP) scan matching. ICP is a well-known scan-to-scan registration method.~\cite{rusinkiewicz2001} and more recently~\cite{pomerleau2015} have surveyed efficient variants of ICP, such as the point-to-plane matching.

~\cite{nuchter2007} give a good review of different 6-DOF LiDAR methods based on 2D or 3D depth sensors, but their solution uses only a stop-scan-go strategy.~\cite{bosse2009} studies a continuous spinning 2D laser. They build a voxel grid from laser points and in each voxel compute shapes to keep only cylindrical and planar areas for matching. Using a 3D laser (Velodyne HDL64),~\cite{moosmann2011} presents a SLAM taking into account the spinning effect of the Velodyne to de-skew the range image along the trajectory. They build a map as a 3D grid structure containing small surfaces, and the de-skewed LiDAR scan is localized with respect to that map. Using a 3D laser,~\cite{ceriani2015} uses a 6-DOF SLAM based on a sparse voxelized representation of the map and a generalization of ICP to find the trajectory.


More recently, LiDAR Odometry And Mapping (LOAM) by~\cite{zhang2017} has become considered state-of-the-art in 6-DOF LiDAR SLAM. They focus on edges and planar features in the LiDAR sweep and keep them in a map for edge-line and planar-planar surface matching.

Different to 2D or 3D spinning LiDAR, RGB-D sensors, such as Kinect, are able to produce dense range images at high frequency. Kinect Fusion~\cite{newcombe2011} presents 3D mapping and localization algorithms using these sensors. They track the 6-DOF position of the Kinect relying on a voxel map storing truncated signed distances to the surface. Such methods are fast and accurate but limited in the volume explored.

\medskip

Our method relies only on 3D LiDAR sensors, such as those produced by Velodyne, and the continuous spinning effects of such sensors. We do not use any data from other sensors, such as IMU, GPS, or cameras. Our algorithm is decomposed in three parts. First, we compute a local de-skewed point cloud from one rotation of the 3D LiDAR. Second, we select specific samples from that point cloud to minimize the distance to the model cloud in the third part. The main contributions of our work are twofold and concern the point selection in each laser scan and the definition of the model map as a point set surface.

\section{Scan egomotion and dynamic object removal}

We define a scan $S$ as the data coming from one rotation of the LiDAR sensor. During the rotation, the vehicle has moved and we need to create a point cloud taking into account that displacement (its egomotion, defined as the movement of the vehicle during the acquisition time of a scan). For that purpose, we assume that the egomotion is relatively similar between two consecutive scans; therfore, we compute the actual egomotion using the previous relative displacement.

We define at any time $t$ the transformation of the vehicle pose as $\tau(t)$ relative to its first position. We only look for discrete solutions for the vehicle positions: $\tau(t_k)$ as the position of the vehicle at the end of the current scan (at time $t_k$ for scan $k$). For any LiDAR measurement at time $t$, the vehicle pose is computed as a linear interpolation between the end of the previous scan $\tau(t_{k-1})$ and the end of the current scan $\tau(t_{k})$.

At time $t_k$, we already know all $\tau(t_{i})$ for $i \leq k-1$ and look for the discrete position $\tau(t_{k})$. To build a local de-skewed point cloud from the current sweep measurements, we must have an estimate $\tilde{\tau}(t_{k})$ of the vehicle position at time $t_k$. We use only previous odometry and define $\tilde{\tau}(t_{k}) = \tau(t_{k-1})*\tau(t_{k-2})^{-1}*\tau(t_{k-1})$.

We build the point cloud scan $S_k$ using a linear interpolation of positions between $\tau(t_{k-1})$ and $\tilde{\tau}(t_{k})$. That egomotion is a good approximation if we assume that the angular and linear velocities of the LiDAR are smooth and continuous over time. Next, we do a rigid registration of that point cloud scan to our map to find $\tau(t_{k})$.

Before matching the scan to the model map, we need to remove all dynamic objects from the scan. This is a very complicated task, requiring a high level of semantic information about the scene to be exact. We perform a small object removal instead of a dynamic object removal and discard from the scene all objects whose size makes it possible to assume that they could be dynamic objects. 
To begin, we detect the ground points in the scan point cloud using a voxel growing similar to that in~\cite{deschaud2010}. We remove these ground points and cluster the remaining points (clusters are points with a distance to the nearest point less than $0.5~m$ in our case). We discard from the scan small group of points; they can represent pedestrians, cars, buses, or trucks. We remove groups of points whose bounding box is less than $14~m$ in $X_v$, less than $14~m$ in $Y_v$, and less than $4~m$ in $Z_v$. ($(X_v,Y_v,Z_v)$ are the axes of a vehicle frame with $X_v$ pointing right, $Y_v$ pointing forward, and $Z_v$ pointing upward). Even removing all these data, we keep enough information about large infrastructure, such as walls, fences, facades, and trees (those with a height of more than $4~m$). Finally, we add back the ground points to the scan point cloud.

\section{Scan sampling strategy}

Once the unwarped point cloud from a scan has been created, we need to select sampling points to do the matching. The classical ICP strategy is to select random samples like in~\cite{rusinkiewicz2001}.~\cite{gelfand2003} gives a more interesting strategy using the covariance matrix of the scan point cloud to find geometric stable sampling. They show that if they select suitable samples, convergence of ICP is possible in cases where it was not possible with random samples. However, their strategy can be slow for the matching part because they may select a large number of samples.

We propose a different sampling inspired by~\cite{gelfand2003}. Instead of using principal axes of the point cloud from the covariance matrix, we keep the axes of the vehicle frame. We define the LiDAR scan point cloud in the vehicle frame with axes ($(X_v,Y_v,Z_v)$). By doing so, most of the planar areas of the scan point cloud (if they are present) are aligned to the $(X_v,Y_v,Z_v)$ axes. For example, ground points provide observability of the translation along $Z_v$. Facades give observability of the translation along $X_v$ and $Y_v$.

First, we need to compute the normals at every point. To do that quickly, we can compute an approximate normal using the spherical range image of the sensor, similar to~\cite{badino2011}. For every point we keep the planar scalar $a_{2D}$ of its neighborhood, as defined by~\cite{demantke2011}: $a_{2D}=(\sigma_2-\sigma_3)/\sigma_1$ where $\sigma_i=\sqrt{\lambda_i}$ and $\lambda_i$ are eigenvalues of the PCA for the normal computation (see \cite{demantke2011} for more details). Second, we compute the nine values for every point $x_i$ in the scan cloud $S_k$: 
\begin{itemize}
\item $a_{2D}^2 (x_i \times \vec{n_i}) \cdot X_v $
\item $-a_{2D}^2 (x_i \times \vec{n_i}) \cdot X_v $
\item $a_{2D}^2 (x_i \times \vec{n_i}) \cdot Y_v $
\item $-a_{2D}^2 (x_i \times \vec{n_i}) \cdot Y_v $
\item $a_{2D}^2 (x_i \times \vec{n_i}) \cdot Z_v $
\item $-a_{2D}^2 (x_i \times \vec{n_i}) \cdot Z_v $
\item $a_{2D}^2 | \vec{n_i} \cdot X_v |$ 
\item $a_{2D}^2 | \vec{n_i} \cdot Y_v |$ 
\item $a_{2D}^2 | \vec{n_i} \cdot Z_v |$ 
\end{itemize}

It is not mandatory in our method to have planar zones in the environment, but such zones allow us to improve the quality of matching compared to non-planar zones; that is why we have $a_{2D}$ in the formulation of choice of samples.
The first 6 values give the contribution of the point $x_i$ of the scan to the observability of the different unknown angles (roll, pitch, yaw) of the vehicle (we see that we provide more important contribution to points far from the sensor center). The 3 last values give the contribution of the point $x_i$ to the observability of the unknown translations (same importance for points far or close to the sensor center). We sort the nine lists in descending order so that the first points of every list are points with more observability in relation to the unknown parameters. During the matching part, we select from each list a sample $x$ starting from the beginning of the list. We find the closest point $p_c$ of $x$ in the model cloud. We keep sample $x$ only if $\|x-p_c\|\leq r$. The parameter $r$ remove outliers between the scan and model cloud. We do this until we find $s$ samples from each list. We may have the same point with a good score in different lists, so we will use it multiple times as sample during the matching process. In any case, we have in total $9s$ samples (we choose the parameter $s$ to keep fewer points than the size of the scan $S_k$).

\begin{figure}[!ht]
\begin{center}
\includegraphics[width=0.92\linewidth]{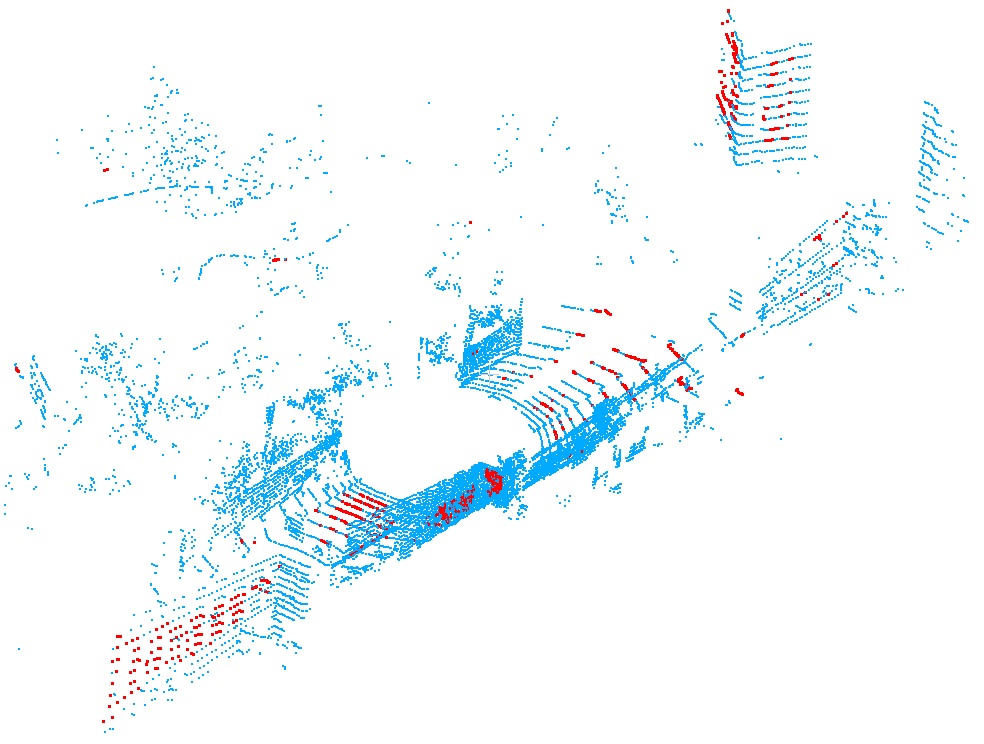}
\end{center}
\caption[width=\linewidth]{Our sampling strategy on a scan point cloud. The points in red are selected samples to do the scan matching. We can see there are points far from the sensor center to better lock the rotations and points on the most planar zones for better matching.}
\label{fig:sampling}
\end{figure}

Figure~\ref{fig:sampling} shows an example of $s=100$ points taken from each list for a scan composed of $13683$ points. For our experiments, we used only $900$ points as samples (around $7\%$ of the scan) to do the matching. It is important to have the minimum number of sampling points (the speed of the matching process depends mainly on that number), but at the same time, we need enough points to have good observability of all parameters of the transformation $\tau(t_k)$. We define as $\tilde{S_k}$ the subset of points in $S_k$ chosen by our sampling strategy.

\section{Scan-to-Model matching with Implicit Moving Least Squares (IMLS) surface representation}

KinectFusion~\cite{newcombe2011} is a well-known SLAM based on the Kinect depth sensor. They do scan-to-model matching using an implicit surface from~\cite{curless1996} as a model. The implicit surface is defined by a Truncated Signed Distance Function (TSDF) and is coded in a voxel map.~\cite{newcombe2011} show great results of scan-to-model matching compared to classical scan-to-scan matching. The problem of TSDF is that the surface is defined by a voxel grid (empty, SDF, unknown) and then is usable only in a small volume space. That is why that TSDF representation cannot be used in large outdoor environments for autonomous driving. In our SLAM, we use the same scan-to-model strategy, but we chose a different surface representation. We take the Implicit Moving Least Square (IMLS) representation computed directly on the map point cloud of the last $n$ localized scans.

Point set surfaces are implicit definitions of surfaces directly on point clouds. In~\cite{levin2004}, Levin is the first to define a Moving Least Square (MLS) surface, the set of stable points of a projection operator. It generates a $C^\infty$ smooth surface from a raw noisy point cloud. Later,~\cite{kolluri2008} defined the IMLS surface: the set of zeros of a function $I(x)$. That function $I(x)$ also behaves as a distance function close to the surface.

We define our point cloud map $P_k$ as the accumulation of $n$ previous localized scans. That point cloud $P_k$ contains noise because of the LiDAR measurements but also errors in localization of the previous scans.

Using the IMLS framework by~\cite{kolluri2008}, we define the function $I^{P_k}(x)$ using equation~\ref{eq:imls} as an approximate distance of any point $x$ in $\mathbb{R}^3$ to the implicit surface defined by the point cloud $P_k$:

\begin{equation}
I^{P_k}(x) = \frac{\sum_{p_i \in P_k}{W_i(x)((x-p_i) \cdot \vec{n_i})}}{\sum_{p_j \in P_k}{W_j(x)}},
\label{eq:imls}
\end{equation}
where $p_i$ are points of the point cloud $P_k$ and $\vec{n_i}$ normals at point $p_i$.

The weights $W_i(x)$ are defined as $W_i(x)=e^{-\|x-p_i\|^2 / h^2}$ for $x$ in $\mathbb{R}^3$. Because the function $W_i(x)$ decreases quickly when points $p_i$ in $P_k$ are far from $x$, we keep only points of $P_k$ inside a ball $B(x,r)$ of radius $r$ (when $r=3h$ and points $p_i$ further than $r$ to $x$, $W_i(x) \leq 0.0002$). The parameter $r$ is the maximum distance for neighbor search and rejected outliers are seen as having no correspondence between the scan and the map (as described in the previous section). $h$ is a parameter for defining the IMLS surface, has been well studied in previous papers~\cite{kolluri2008} and depends on the sampling density and noise of the point cloud $P_k$.

We want to localize the current scan $S_k$ in point cloud $P_k$. To do so, we want to find the transformation $R$ and $t$ that minimizes the sum of squared IMLS distances: $\sum_{x_j \in \tilde{S_k}} {|I^{P_k}(R x_j + t)|^2}$. Due to exponential weights, we cannot approximate that nonlinear least-square optimization problem by a linear least-square one, as in ICP point to plane. Instead of minimizing that sum, we project every point $x_j$ of $\tilde{S_k}$ on the IMLS surface defined by $P_k$: $y_j = x_j - I^{P_k}(x_j)\vec{n_j}$ where $\vec{n_j}$ is the normal of the closest point $p_c$ to $x_j$ and is a good approximation of the surface normal at the projected point $y_j$.

Now, we have a point cloud $Y_k$, the set of projected points $y_j$ and we look for the transformation $R$ and $t$ that minimizes the sum $\sum_{x_j \in \tilde{S_k}}{| \vec{n_j} \cdot (R x_j + t - y_j) |^2 }$. Like in ICP point-to-plane, we can now make the small angle assumption on $R$ to get a linear least-square optimization problem that can be solved efficiently (more details in the technical report~\cite{low2004}). We compute $R$ and $t$ and move the scan $S_k$ using that transformation. The process is then started again: project the points $x_j$ of the scan on the IMLS surface to form $Y_k$, find the new transformation $R$ and $t$ between the scan $S_k$ and point cloud $Y_k$ and move the scan with the found transformation $R$ and $t$. We iterate until a maximum number of iterations has been made. The final transformation $\tau(t_k)$ is the composition of the transformation between the first and last iteration of the scan during the matching process and the estimate position $\tilde{\tau}(t_k)$. Now, we can compute a new point cloud from raw data of the current scan by linear interpolation of vehicle position between $\tau(t_{k-1})$ and $\tau(t_k)$. We add that point cloud to the map point cloud and remove the oldest point cloud scan to always keep $n$ scans in the map. 

With IMLS formulation, we need to compute normal $\vec{n_j}$ from the point cloud $P_k$ for every query point $x_j$ of scan $S_k$. This is done at every iteration during the neighbor search but only for the selected samples using the same normal for neighbors points (so $9s$ normals are calculated at each iteration).

\begin{figure}[!ht]
\begin{center}
\includegraphics[width=0.92\linewidth]{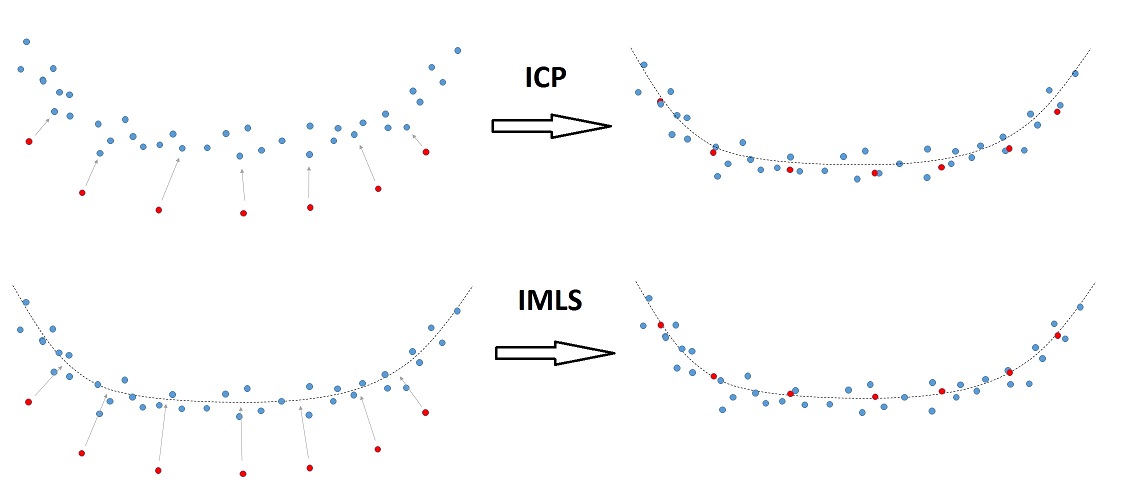}
\end{center}
\caption[width=\linewidth]{Schematic example of classical ICP scan matching compared to our IMLS scan-to-model framework. The blue points are the noisy point cloud $P_k$ from the $n$ previous localized scans. The red points are points from the new scan. The first row shows an example of the first and last iteration of ICP point-to-plane matching. At each iteration, we minimize the distance to the closest point. The second row shows an example of the first and last iteration of our IMLS point-to-model distance. The dashed black line is the IMLS surface. At each iteration, we minimize the sum of the distance to the IMLS surface (for simplification purposes, we removed the normal formulation from the schema).}
\label{fig:imls}
\end{figure}

Figure~\ref{fig:imls} is a schematic example of the difference between our IMLS scan-to-model matching and a classical ICP scan-to-point cloud matching. The advantage of this formulation is to move the scan converge towards the implicit surface and improve the quality of matching.

\section{Experiments}

Our IMLS SLAM has been implemented in C++ using only the FLANN library (for nearest neighbor research with k-d tree) and Eigen. We have done tests on a real outdoor dataset from LiDAR Velodyne HDL32 and Velodyne HDL64 spinning at 10~Hz (each scan has been acquired during $100~ms$). The method runs on one CPU core at 4~GHz and uses less than 1~Go of RAM. Velodyne HDL32 and HDL64 are rotating 3D LiDAR sensors with 32 and 64 laser beams.

For all experiments, we used $s=100$ for the number of sampling points in each list, $h=0.06~m$ (for the IMLS surface definition), $r=0.20~m$ (maximum distance for neighbors search), $20$ is the number of matching iterations (to keep a constant timing process instead of using convergence criteria), and $n=100$ is the number of scans we have in the model point cloud.

\subsection{Tests on our Velodyne HDL32 dataset}

\begin{figure}[!ht]
\begin{center}
\includegraphics[width=0.92\linewidth]{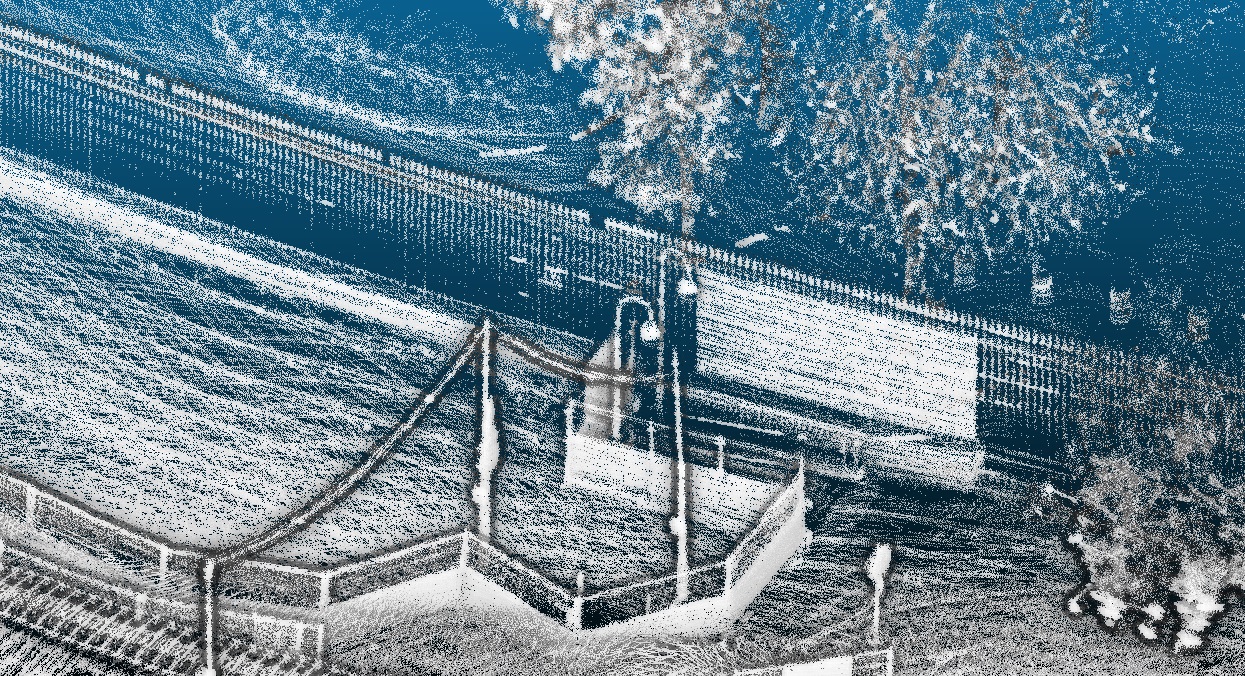}
\end{center}
\caption[width=\linewidth]{Small portion of the point cloud generated by our IMLS SLAM (red rectangle in Fig.~\ref{fig:traj_velo32_paris}). The visible details of the fence show we get a good assumption on the egomotion for each laser scan.}
\label{fig:pc_velo32_paris}
\end{figure}

To test the SLAM, we made a $4~km$ acquisition in the center of Paris with a Velodyne HDL32 sensor in a vertical position on the roof of a vehicle (total of 12951 scans). This is a $2~km$ loop we did two times and came back exactly at the same place (less than a meter difference). We then measured the distance between the first and last localized scan as an error metric. Figure~\ref{fig:traj_velo32_paris} shows the trajectory of our IMLS SLAM and the trajectory of a classical ICP scan-to-scan matching (equivalent to $n=1$). We can see the good superposition of the two loops with our SLAM. Figure~\ref{fig:pc_velo32_paris} shows a small portion of the point cloud generated from the SLAM. We can see fine details like the fence. This means we have good egomotion of the vehicle during each scan. The distance error between the first and last scan with our IMLS SLAM is $16~m$, a drift of only 0.40\%.

\begin{figure}[!ht]
\begin{center}
\includegraphics[width=0.92\linewidth]{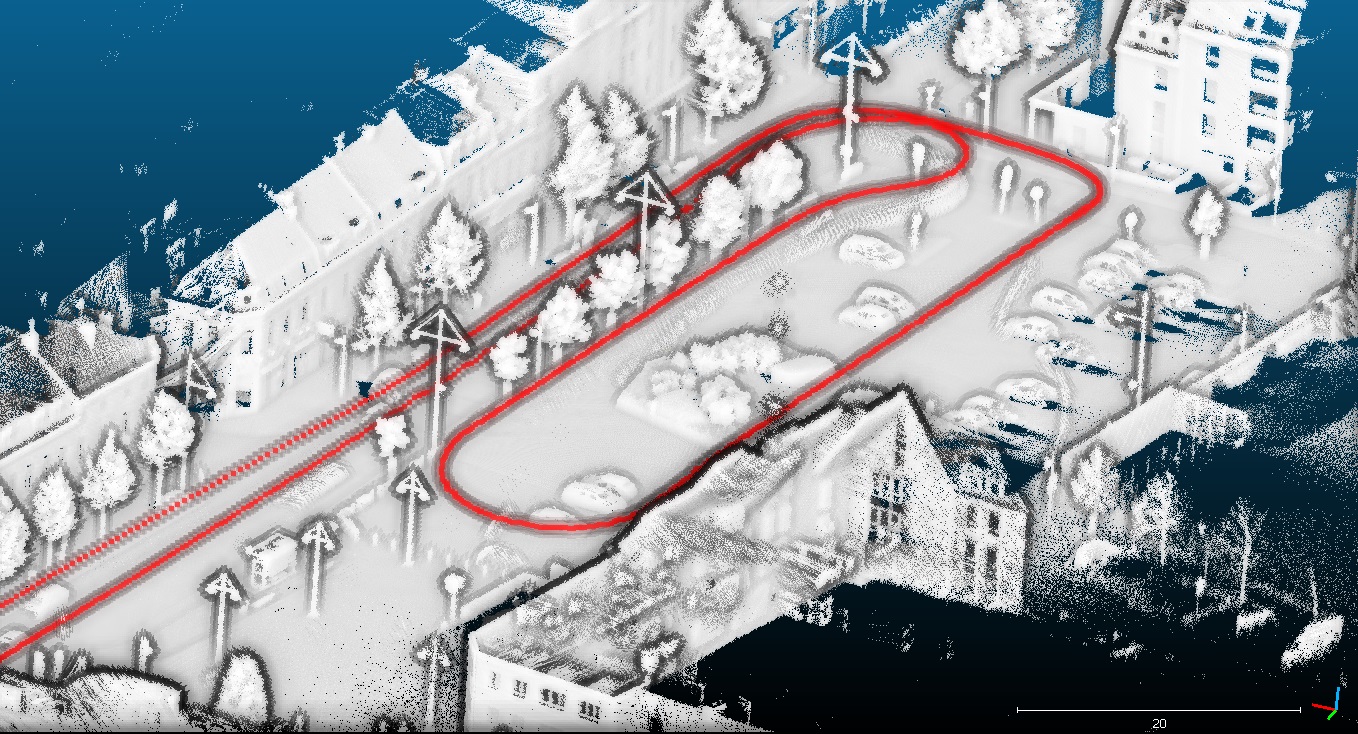}
\end{center}
\caption[width=\linewidth]{Trajectory in red and point cloud of our IMLS SLAM in the city of Lille. We can see the quality of the map with no duplicate objects despite numerous turns.}
\label{fig:pc_traj_velo32_lille}
\end{figure}

We tested our SLAM with the Velodyne HDL32 in a different orientation. The Velodyne is still on the roof of a vehicle but is tilted 60 degrees in pitch. The acquisition has been made in a square of the city of Lille with many turns to test the robustness of the matching (total of $1500$ scans). Figure~\ref{fig:pc_traj_velo32_lille} shows in red the trajectory of the vehicle computed by our SLAM and the generated point cloud. We can see that there are no duplicate objects despite having done many turns in the square. The point cloud provides a qualitative evaluation of the mapping.

\subsection{Tests on the public dataset KITTI with Velodyne HDL64}

We tested our SLAM method on the public dataset KITTI. The odometry evaluation dataset has 22 sequences with stereo and LiDAR data (results of 86 methods are available online). The LiDAR is a vertical Velodyne HDL64 on the roof of a car. Eleven sequences are provided with ground truth (GPS+IMU navigation) and 11 sequences are given without ground truth for odometry evaluation. The dataset is composed of a wide variety of environments (urban city, rural road, highways, roads with a lot of vegetation, low or high traffic, etc.). More details are available at~\footnote{\url{http://www.cvlibs.net/datasets/kitti/eval_odometry.php}} or in~\cite{geiger2012} regarding the metric used for evaluation. The LiDAR scans are de-skewed with an external odometry, so we did not apply our egomotion algorithm to this dataset.

In the training dataset, we get 0.55\% drift in translation and 0.0015 deg/m error in rotation. We can compare the results to~\cite{ceriani2015}, who had around 1.5\% drift error in translation and 0.01 deg/m error in rotation. Table~\ref{table:training_kitti_results} compares our results for the training dataset to LOAM~\cite{zhang2017}. We see we outperfom previously published results.


\begin{table}[h]
\begin{center}
\begin{tabular}{|c||c|c|c|}
\hline
\textbf{Sequence} & \textbf{Environment} & \textbf{LOAM~\cite{zhang2017}} & \textbf{Our SLAM}\\
\hline
0 & Urban & 0.78\% & \textbf{0.50\%}\\
\hline
1 & Highway & 1.43\% & \textbf{0.82\%}\\
\hline
2 & Urban+Country & 0.92\% & \textbf{0.53\%}\\
\hline
3 & Country & 0.86\% & \textbf{0.68\%}\\
\hline
4 & Country & 0.71\% & \textbf{0.33\%}\\
\hline
5 & Urban & 0.57\% & \textbf{0.32\%}\\
\hline
6 & Urban & 0.65\% & \textbf{0.33\%}\\
\hline
7 & Urban & 0.63\% & \textbf{0.33\%}\\
\hline
8 & Urban+Country & 1.12\% & \textbf{0.80\%}\\
\hline
9 & Urban+Country & 0.77\% & \textbf{0.55\%}\\
\hline
10 & Urban+Country & 0.79\% & \textbf{0.53\%}\\
\hline
\end{tabular}
\end{center}
\caption{Comparison of drift KITTI training dataset between our SLAM and LOAM state-of-the-art odometry (LOAM results are taken from paper~\cite{zhang2017}). We can see we outperform on every type of environment.}
\label{table:training_kitti_results}
\end{table}

In the test dataset, we have 0.69\% drift in translation and 0.0018 deg/m error in rotation (visible on the KITTI website). It is better than the state-of-the-art published results of LOAM in~\cite{zhang2017} where they had 0.88\% drift. On the KITTI website, LOAM improved their results, which were a little better than ours with 0.64\% drift. 

The drift we get on the KITTI benchmark is not as good as the results we obtained with the Velodyne HDL32. This is due to three facts. First, we found a distortion of the scan point clouds because of a bad intrinsic calibration (we did a calibration of the intrinsic vertical angle of all laser beams of $0.22$ degrees using the training data). Second, we found big errors in GPS data (used as ground truth) with, for example, more than $5~m$ in the beginning of sequence~8. Third, the environment has more variety (vegetation, highways, etc.) than the urban environment for the Velodyne HDL32 test.

We measure the contributions of the different parts of the algorithm to the KITTI training dataset. Table~\ref{table:object_removal_kitti} shows the importance of dynamic object removal. Table~\ref{table:sampling_kitti} shows the contribution of our sampling strategy compared to random sampling and geometric stable sampling (from~\cite{gelfand2003}). Table~\ref{table:model_kitti} shows the importance of parameter $n$. We keep improving the results by taking $n=100$ instead of only $n=10$ scans of the map. We also tried to keep more than $n=100$ scans, but this does not change the results because then the oldest scans are too far from the current scan to have an influence. We also tested changing the parameter $s$ of the number of samples in Table~\ref{table:samples_kitti}. When the number of samples is too small ($s=10$), we do not have enough points to have good observability for matching the scan to the map point cloud. But when the number of samples is too big ($s=1000$), the results are worse because we keep too many points from the scan (as explained in~\cite{gelfand2003}, keeping too many points can alter the constraints to find the final pose).

\begin{table}[h]
\begin{center}
\begin{tabular}{|c||c|}
\hline
\textbf{Object Removal} & \textbf{Drift on KITTI training dataset}\\
\hline
Without & 0.58\%\\
\hline
With & \textbf{0.55}\%\\
\hline
\end{tabular}
\end{center}
\caption{Importance of dynamic object removal of a scan on the KITTI training dataset}
\label{table:object_removal_kitti}
\end{table}

\begin{table}[h]
\begin{center}
\begin{tabular}{|c||c|}
\hline
\textbf{Sampling strategy} & \textbf{Drift on KITTI training dataset}\\
\hline
Random sampling & 0.64\%\\
\hline
Geometric stable sampling~\cite{gelfand2003} & 0.57\%\\
\hline
Our sampling & \textbf{0.55\%}\\
\hline
\end{tabular}
\end{center}
\caption{Importance of sampling strategy of a scan on the KITTI training dataset}
\label{table:sampling_kitti}
\end{table}

\begin{table}[h]
\begin{center}
\begin{tabular}{|c||c|}
\hline
\textbf{Parameter $n$} & \textbf{Drift on KITTI training dataset}\\
\hline
$n=1$ scan & 1.41\%\\
\hline
$n=5$ scans & 0.58\%\\
\hline
$n=10$ scans & 0.56\%\\
\hline
$n=100$ scans & \textbf{0.55\%}\\
\hline
\end{tabular}
\end{center}
\caption{Importance of the parameter $n$, last scans kept as model on the KITTI training dataset}
\label{table:model_kitti}
\end{table}

\begin{table}[h]
\begin{center}
\begin{tabular}{|c||c|}
\hline
\textbf{Parameter $s$} & \textbf{Drift on KITTI training dataset}\\
\hline
$s=10$ samples/list & 0.79\%\\
\hline
$s=100$ samples/list & \textbf{0.55\%}\\
\hline
$s=1000$ samples/list & 0.57\%\\
\hline
\end{tabular}
\end{center}
\caption{Importance of the parameter $s$, the number of samples for the matching part on the KITTI training dataset}
\label{table:samples_kitti}
\end{table}

\begin{figure}[!ht]
\begin{tabular}{c}
	\includegraphics[width=0.92\linewidth]{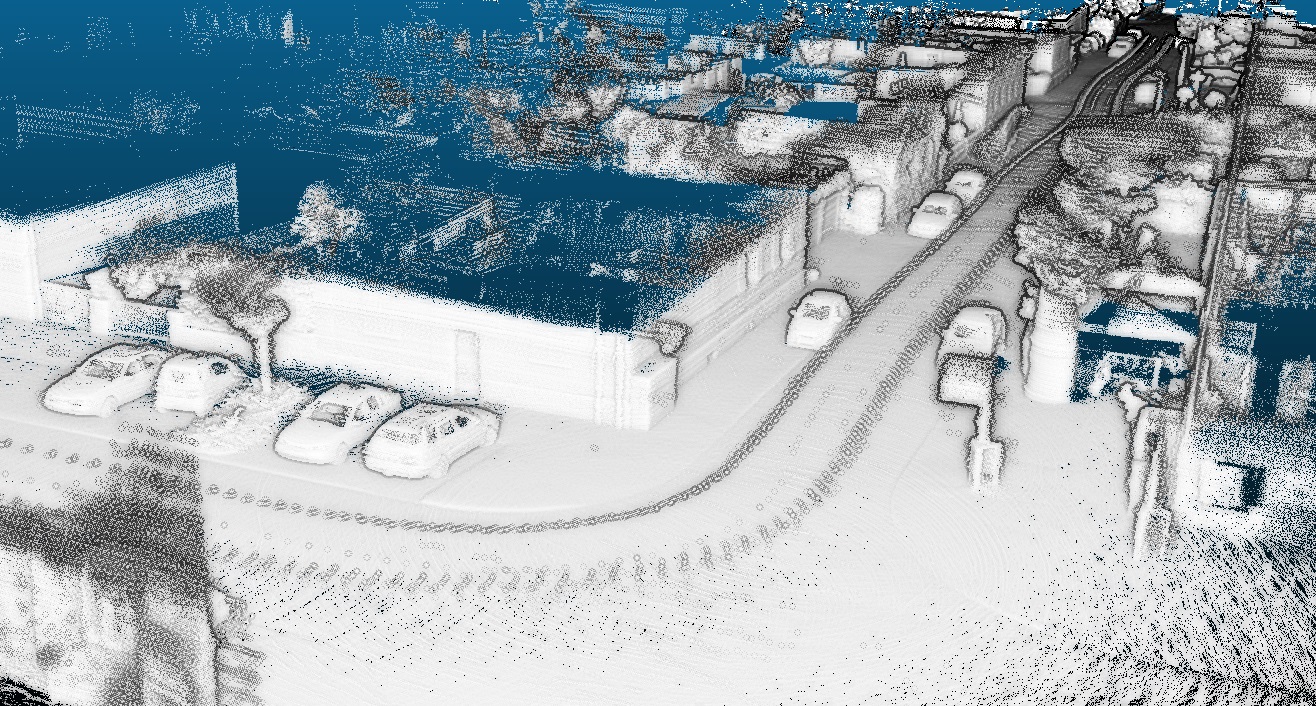} \\
	\includegraphics[width=0.92\linewidth]{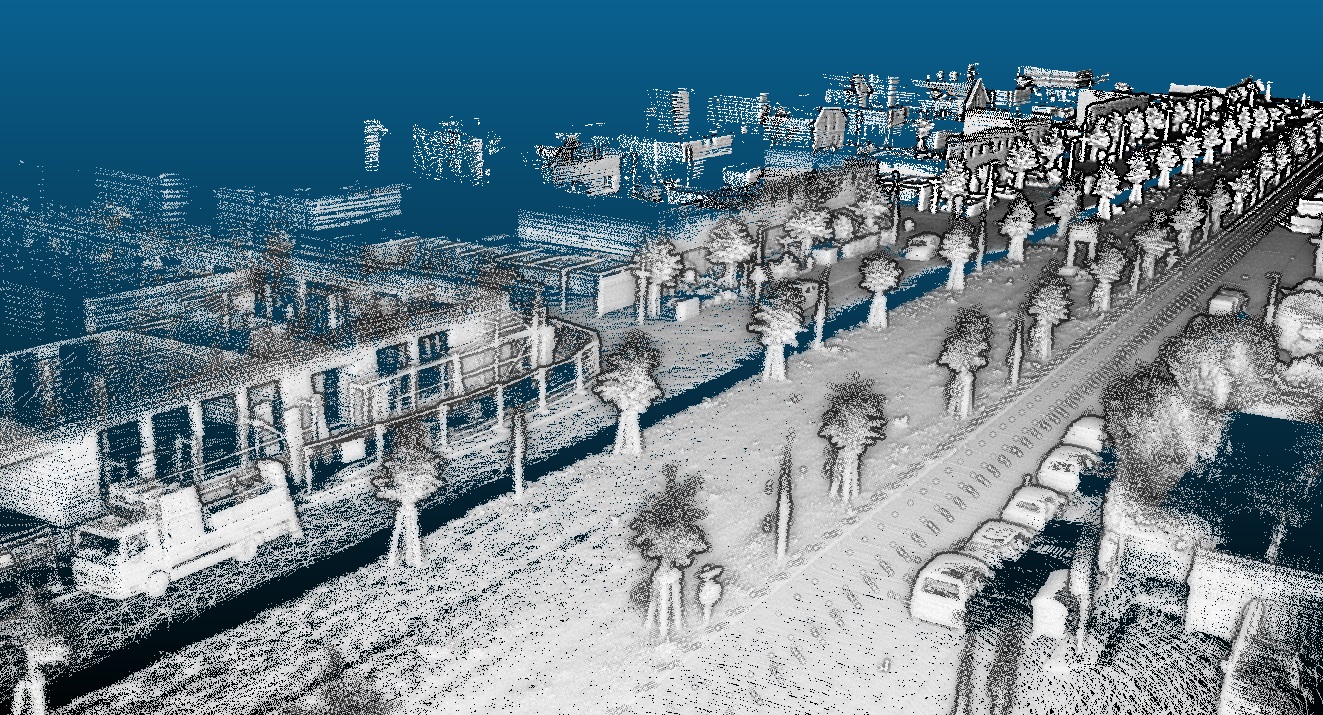} \\
\end{tabular}
\caption[width=\linewidth]{Point clouds generated from sequence 0 (top) and 6 (bottom) of the KITTI training dataset with our LiDAR odometry. We can see the details of the different objects even with multiple passages, like in sequence 6 (bottom).}
\label{fig:kitti_sequence}
\end{figure}

Figure~\ref{fig:kitti_sequence} shows two point clouds produced by our LiDAR odometry in the KITTI dataset. We can see the details of the environment (cars, poles) and the large number of outliers (for which we are robust).

\subsection{Discussion of the processing time}

Our IMLS SLAM implementation is not in real time. First, we compute the normal at every scan using the 3D point cloud of the scan instead of using the fast normal computation of~\cite{badino2011} in the spherical range image (being 17 times faster). This is because the KITTI dataset provides only 3D point cloud and not the raw LiDAR data. It takes $0.2~s$ per scan to do our normal computation. Second, at every scan, we compute a new k-d tree from the whole point cloud $P_k$ to find the nearest neighbors in the IMLS formulation. It takes time, depending on the number $n$ of last scans stored. When $n=100$, it takes $1~s$ per sweep. One solution would be to build a specific k-d tree (keeping the k-d tree between scans by only removing points from the oldest scan and adding points from the previous scan). The matching iterations are very fast (thanks to the limited number of queries with our sampling strategy) and takes $0.05~s$ per scan. So, because of the KITTI dataset and our implementation, our SLAM runs at $1.25~s$ per scan. We think it could be improved with better normal computation and a specific k-d tree to run in real time. For comparison, as explained in~\cite{zhang2017}, LOAM runs at $1~s$ per scan on the KITTI dataset.

\section{Conclusion}

We presented a new 3D LiDAR SLAM based on a specific sampling strategy and new scan-to-model matching. Experiments showed low drift results on Velodyne HDL32 dataset and among best results on KITTI benchmark. We think that our method could be improved to run in real time in future work.


{\small
\bibliographystyle{IEEEtran}
\bibliography{IEEEabrv,mybib}
}

\end{document}